\documentclass{article}

\usepackage{microtype}
\usepackage{graphicx}
\usepackage{subfigure}
\usepackage{booktabs} % for professional tables

\usepackage{hyperref}

\usepackage[accepted]{icml2024}

\usepackage{amsmath}
\usepackage{amssymb}
\usepackage{mathtools}
\usepackage{amsthm}
\usepackage{xcolor}

\usepackage[capitalize,noabbrev]{cleveref}

\theoremstyle{plain}

\theoremstyle{definition}

\theoremstyle{remark}

\usepackage[textsize=tiny]{todonotes}

\def\etal{\emph{et al.}}

\DeclareUnicodeCharacter{2212}{-}
\newlength\savewidth
\usepackage{makecell}
\usepackage{multirow}
\usepackage{pifont}
\newcommand{\cmark}{\ding{51}}%
\newcommand{\xmark}{\ding{55}}%

\begin{document}

\twocolumn[
\icmltitle{Data-efficient Large Vision Models through Sequential Autoregression}



\icmlsetsymbol{equal}{*}

\begin{icmlauthorlist}
\icmlauthor{Jianyuan Guo}{equal,usyd}
\icmlauthor{Zhiwei Hao}{equal,sch}
\icmlauthor{Chengcheng Wang}{equal,comp}
\icmlauthor{Yehui Tang}{comp}
\icmlauthor{Han Wu}{usyd}
\icmlauthor{Han Hu}{sch}
\icmlauthor{Kai Han}{comp}
\icmlauthor{Chang Xu}{usyd}
\end{icmlauthorlist}

\begin{center}
jianyuan\_guo@outlook.com; \{haozhw,hhu\}@bit.edu.cn; \{wangchengcheng11,yehui.tang\}@huawei.com;
\end{center}

\icmlaffiliation{usyd}{School of Computer Science, Faculty of Engineering, University of Sydney, Sydney, Australia}
\icmlaffiliation{comp}{Huawei Noah's Ark Lab, Beijing, China}
\icmlaffiliation{sch}{School of information and Electronics, Beijing Institute of Technology, Beijing, China}

\icmlcorrespondingauthor{Kai Han}{kai.han@huawei.com}
\icmlcorrespondingauthor{Chang Xu}{c.xu@sydney.edu.au}

\icmlkeywords{Machine Learning, ICML}

\vskip 0.3in
]



\printAffiliationsAndNotice{\icmlEqualContribution} % otherwise use the standard text.

\begin{abstract}
Training general-purpose vision models on purely sequential visual data, eschewing linguistic inputs, has heralded a new frontier in visual understanding. These models are intended to not only comprehend but also seamlessly transit to out-of-domain tasks.
However, current endeavors are hamstrung by an over-reliance on colossal models, exemplified by models with upwards of 3B parameters, and the necessity for an extensive corpus of visual data, often comprising a staggering 400B tokens~\cite{lvm}. 
In this paper, we delve into the development of an efficient, autoregression-based vision model, innovatively architected to operate on a limited dataset. We meticulously demonstrate how this model achieves proficiency in a spectrum of visual tasks spanning both high-level and low-level semantic understanding during the testing phase. Our empirical evaluations underscore the model's agility in adapting to various tasks, heralding a significant reduction in the parameter footprint, and a marked decrease in training data requirements, thereby paving the way for more sustainable and accessible advancements in the field of generalist vision models. The code is available at \href{https://github.com/ggjy/DeLVM}{https://github.com/ggjy/DeLVM}.

\end{abstract}

\section{Introduction}
Training a generalist model capable of executing diverse tasks simultaneously—and with the agility to tackle new tasks given just few examples—represents a pivotal stride toward artificial general intelligence within the computer vision community. In the realm of contemporary natural language processing (NLP), large language models trained via autoregression, such as GPT~\cite{gpt3,gpt4}, have demonstrated the remarkable ability to comprehend and generate natural language text, particularly excelling in complex and nuanced contexts. These models leverage language sequences as a universal interface, facilitating rapid adaptation to a variety of language-centered tasks with minimal prompting and examples.

However, the landscape of computer vision is markedly different. Unlike the uniformity of input-output structures in language tasks, visual tasks exhibit a rich diversity in their formats, posing significant challenges to the development of comparable generalist models. Recently, advances in LVM~\cite{lvm} have redefined `visual sentences', enabling the representation of both raw images and annotated data without requiring meta-knowledge beyond the pixel level. This paradigm shift opens a new vista for the evolution of generalist models within the visual domain.

\begin{figure*}[t]
\centering
\includegraphics[width=1\linewidth]{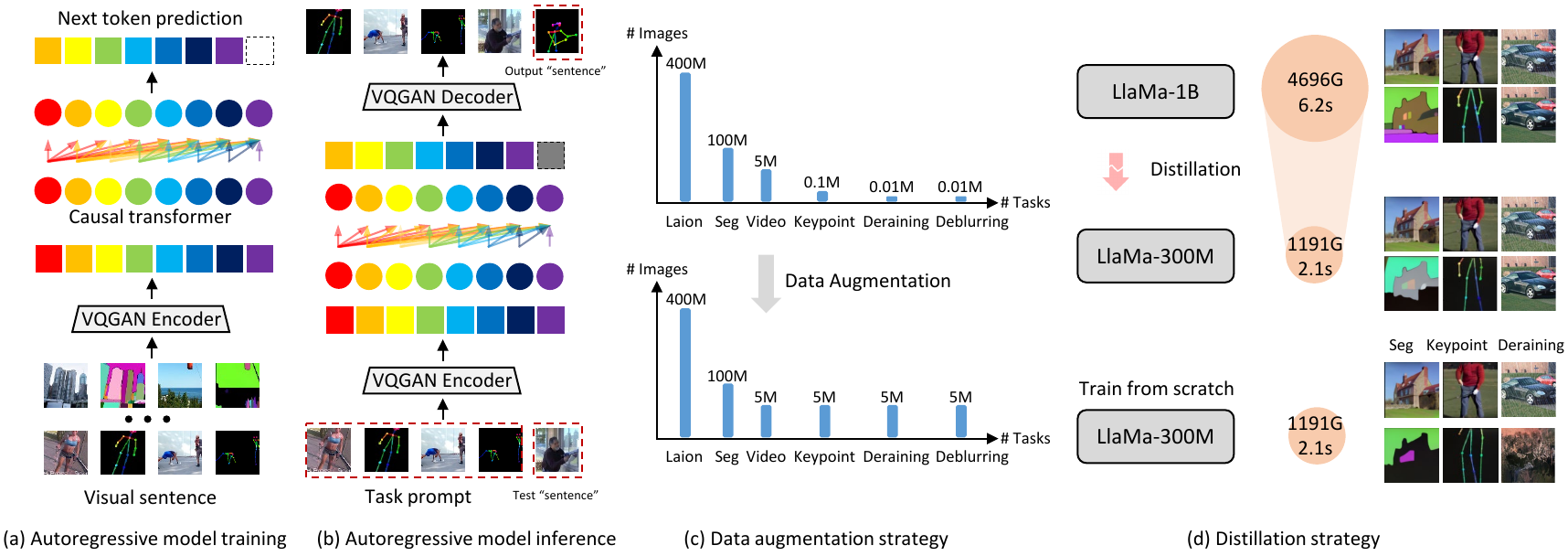}
\vspace{-20pt}
\caption{\small{An overview of our framework. We follow the autoregressive setting proposed in LVM~\cite{lvm}, which encodes input images into a 1D sequence. We further explore the data-efficient LVMs via data augmentation and knowledge distillation strategy.}}
\vspace{-10pt}
\label{fig:framework}
\end{figure*}

Nevertheless, while the capability of LVM is bolstered by the employment of large-scale datasets, such as 1.64 billion images, this reliance simultaneously engenders complexities in the training process of LVM. One obstacle arises from the imbalance of datasets across different tasks.
Our visual world naturally exhibits a long-tailed distribution of different tasks.
The model performance on tasks with limited data representation is substantially impaired when there is an uneven distribution of data across tasks, with certain tasks having an abundance of data and others suffering from a deficiency.
For example, the segmentation benchmark SA-1B~\cite{sam}, containing 11 million images, provides a wealth of data compared to the keypoint detection benchmark COCO~\cite{coco}, containing only 0.2 million images. Direct training on the combined dataset of these two benchmarks results in the model's inability to learn keypoint detection due to the overwhelming amount of segmentation data, as iilustrated in Figure~\ref{fig:framework}.
This observation underpins our simple data augmentation strategy that automatically enrich the dataset of relatively small size and achieve equilibrium among disparate tasks. Specifically,we enhance the tail datasets by randomly augmenting the training samples, which we find yield stronger performance compared with traditional re-sampling strategies.

The significant performance gain brought by the proposed data augmentation compels us to contemplate: Given that a substantial reduction in the volume of data required for training does not adversely affect model efficacy, could there potentially exist superfluousness within the model's parameterization? We resort to knowledge distillation (KD) methods to enhance the performance of compact LVMs. In our study, we demonstrate the effectiveness of KD in autoregressive LVMs. While models with larger capacity initially exhibit superior performance and surpasses the student model trained from scratch by a substantial margin, the incorporation of KD significantly narrows the performance gap and shed light on building efficient LVMs.

In general, this paper explores the development of data-efficient autoregressive large vision models (LVMs). Our study focuses on data augmentation strategy, especially in scenarios with long-tail distributions across different tasks. We demonstrate that simple data augmentation techniques yield satisfactory results compared to re-sampling baselines. We also leverage KD to create more compact and efficient LVMs, which leads to a significant reduction in validation loss, with improvement in accuracy and decrease in perplexity. This highlights the potential of KD to bridge the performance gap and enhance the capabilities of autoregressive LVMs. Our work aims to advance the understanding and construction of more efficient autoregressive vision models capable of simultaneously addressing various vision tasks.

\section{Autoregressive Large Vision Models}

Inspired by the achievements of LLMs~\cite{gpt3,llama2,palm}, various endeavors have been made to develop autoregressive models specifically tailored for vision tasks.
An outstanding breakthrough in this domain, referred to as the LVM~\cite{lvm}, transforms visual data into \emph{visual sentences}, thereby enabling the uniform modeling of diverse vision tasks.
A typical LVM consists of two key modules: (1) a VQGAN~\cite{vqgan} for input image tokenization; (2) a transformer model~\cite{llama} for sequential autoregressive prediction.

\textbf{VQGAN tokenizer.}
The initial step of modeling images via a transformer-based autoregressive approach involves converting original images into discrete tokens.
A VQGAN~\cite{vqgan} is utilized for this tokenization process, composed of three main components: an encoder, a decoder, and a trainable codebook.
When training a VQGAN model, the encoder takes an image as input and projects it into a grid of features, which are then mapped to specific codes in the codebook.
The decoder is used to reconstruct the original image based on the grid of codes.
A pretrained VQGAN model is capable of converting the input image into a sequence of tokens following the scan-line order, which can be modeled by an autoregressive model.

\textbf{Autoregressive model.}
After tokenization, visual sentences are constructed by concatenating tokens from multiple images and fed into a causal transformer model which is employed for autoregressive prediction.
The goal of the causal transformer is to predict the next token at each position, using cross-entropy as the loss function.
Specifically, considering an input sentence of length $L$ as $s_i=\{x_1, x_2, ...,x_{L-1}, x_L\}$, where $x_l$ represents the $l$-th token, the objective of the causal transformer is to predict $s_o=\{x_2, x_3, ..., x_{L}, \varnothing\}$.

\textbf{Prompted inference.}
After training, LVM is capable of performing inference on downstream tasks via vision prompting.
Firstly, several input-output image pairs are converted into a visual sentence, which serves as a task definition.
This sentence is then combined with the test image, which is also tokenized via the VQGAN encoder.
Based on the concatenated sentence, the model generates output tokens based on the input sequential tokens.
The final predicted image is then obtained by decoding above generated tokens using the decoder of the VQGAN model.

\begin{figure}[!t]
\vspace{-2pt}
\begin{center}
\centerline{\includegraphics[width=0.98\columnwidth]{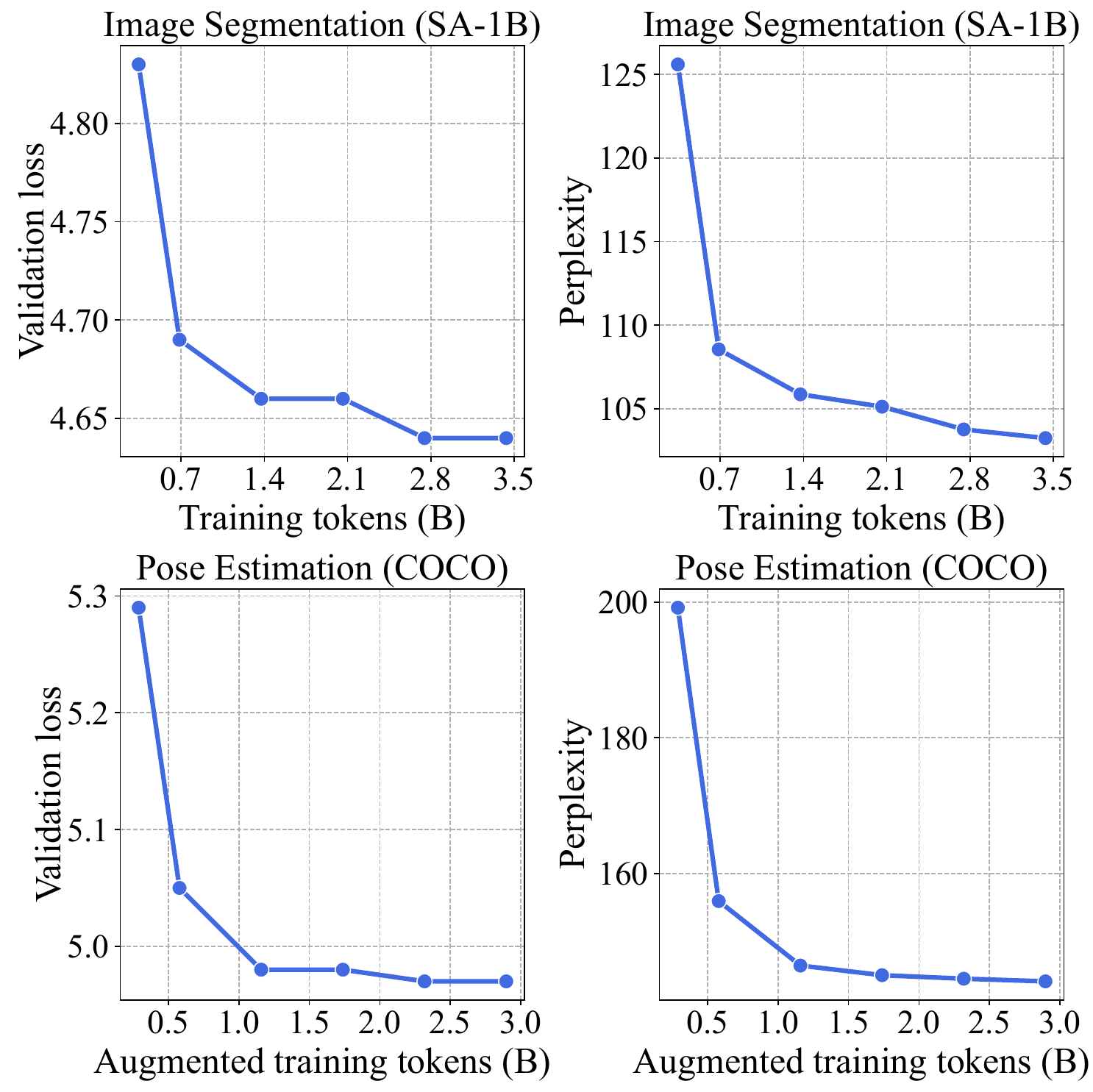}}
\caption{\textbf{Data augmentation yields a similar effect to the introduction of new data.} We train a LLaMA-300M model on subsets of SA-1B and augmented COCO-Pose datasets with a fixed 12K training steps. In segmentation task with ample available data, enhanced model performance is observed upon introducing more training data. Conversely, for human pose estimation task with limited data, augmenting the dataset has a comparable impact to the introduction of new training data. The original COCO-Pose dataset contains approximately 0.03 billion tokens.}
\label{fig:data_aug_loss_and_ppl}
\end{center}
\vspace{-2pt}
\end{figure}

\section{Data Augmentation}
In common practices of training highly proficient LLMs, the training duration is often restricted to just one epoch to mitigate the risk of overfitting~\cite{gpt3,llama,llama2}.
In contrast to the NLP field, where numerous large-scale corpora are readily available, many computer vision tasks encounter a scarcity of extensive datasets.
The scarcity makes it impractical to directly transfer the training schedule settings from language tasks to vision tasks.
Fortunately, various data augmentation techniques commonly employed in traditional computer vision tasks can be leveraged to enhance the training data for LVMs.
We initiate the examination of data augmentation effectiveness by training LVM on various vision tasks.

\subsection{Single Task}
We initiate our study by conducting experiments on a single task.
It is intuitive that the introduction of new training data enhances overall model performance.
To delve deeper into this effect, we proceed to train the model on an image segmentation task using varying amounts of available data.
Following the configuration of LVM~\cite{lvm}, we employ a causal LLaMA model~\cite{llama} with 300M parameters for autoregressive modeling.
For tokenization, we utilize an off-the-shelf VQGAN trained by Chang \etal~\cite{muse}, which translates each image into 256 discrete tokens.
The VQGAN has a codebook size of 8192, with each code in the codebook having a dimension of 64. To align the dimension of the code with that of the transformer model, we insert a learnable linear layer between these two modules.
Regarding the dataset, we utilize various subsets of the SA-1B~\cite{sam}, spanning from 1\% to 10\%.
The training steps for all models remain fixed at 12K.
Cross-entropy loss and perplexity on a withheld subset of SA-1B (equivalent to 1\% of the dataset) serve as the metrics.
The top two figures in Figure~\ref{fig:data_aug_loss_and_ppl} illustrate the impact of available data on model performance.
Obviously, as more new data is introduced during training, there is a significant increase in model performance.
For instance, when the utilized data from SA-1B increases from 1\% to 10\% (0.34B tokens to 3.43B tokens), the validation loss decreases by 0.19, and perplexity decreases by 22.4.

However, in certain tasks, such as human pose estimation,  the availability of data is constrained, making it challenging to enhance model performance by introducing new samples.
To bridge this gap, we employ data augmentation techniques to augment the existing dataset.
Specifically, for each sample in the dataset, we apply random crop and random flip operations to generate several augmented versions.
These augmented samples are then tokenized by the VQGAN model and incorporated into training alongside the tokenized original dataset.
We assess the impact of augmentation by varying the augmentation range from 10 to 100 times.
As depicted in the bottom two figures in Figure~\ref{fig:data_aug_loss_and_ppl}, expanding the data scale through data augmentation also contributes to improved model performance.
The decreasing trend observed in validation loss and perplexity as the augmentation times increase mirrors the pattern observed when introducing new samples.
This outcome underscores that data augmentation has a comparable effect to acquiring more new data, offering a straightforward approach to effectively train LVMs in data-limited scenarios.

\begin{table*}[!ht]
  \renewcommand\tabcolsep{4pt}
  \caption{\textbf{Balancing datasets across multiple tasks through augmentation enhances performance.} We train a LLaMA-300M on a mixed dataset involving three tasks. Three configurations for handling the unbalanced dataset are compared. Compared to directly training on the unbalanced dataset, mitigating the long-tail distribution through re-sampling led to inferior results. Conversely, achieving balance through dataset augmentation yielded the best overall performance. ``-'' indicates collapsed results (larger than $10^8$).}
  \label{tab:3task_aug}
  \small
  \begin{center}
    \begin{tabular}{l|ccc|ccc|ccc}
      \toprule
      & \multicolumn{3}{c|}{\textbf{Image Segmentation}} & \multicolumn{3}{c|}{\textbf{Pose Estimation}} & \multicolumn{3}{c}{\textbf{Image Deraining}}  \\
      & \multicolumn{3}{c|}{(10\% of SA-1B: 3.434B tokens)} & \multicolumn{3}{c|}{(COCO-Pose: 0.029B tokens)} & \multicolumn{3}{c}{(Rain13K: 0.007B tokens)}  \\
      \textbf{Dataset configuration} & loss $\downarrow$ & accuracy $\uparrow$ & perplexity $\downarrow$ & loss $\downarrow$ & accuracy $\uparrow$ & perplexity $\downarrow$ & loss $\downarrow$ & accuracy $\uparrow$ & perplexity $\downarrow$ \\
      \midrule
      Unbalanced                     & 4.48              & 20.32               & \enspace88.59           & \enspace4.95      & 19.86               & 141.91                  & \enspace5.69      & 11.89               & 279.88                  \\
      Balanced by re-sampling        & 5.01              & 16.68               & 151.66                  & 16.11             & 15.75               & -                       & 21.73             & \enspace4.53        & -                       \\
      Balanced by augmentation       & 4.68              & 18.79               & 107.77                  & \enspace4.94      & 20.24               & 140.39                  & \enspace5.64      & 11.57               & 269.04                  \\
      \bottomrule
    \end{tabular}
    \vspace{-2pt}
  \end{center}
\end{table*}

\begin{figure*}[!ht]
\vspace{-4pt}
\begin{center}
\centerline{\includegraphics[width=\linewidth]{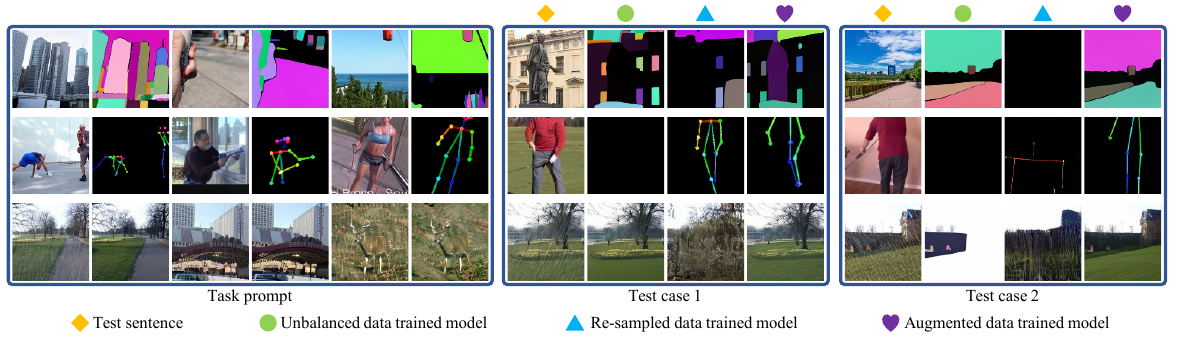}}
\caption{Visualization of inference results generated by models trained on datasets with different balancing schemes. The model trained on an unbalanced dataset exhibits biased performance, excelling primarily in the image segmentation task. Rectifying this data imbalance through re-sampling have unfortunately led to a decrease in performance. Conversely, employing augmentation to balance the dataset results in improved visualization outcomes.}
\label{fig:3tasks_visualization}
\end{center}
\vspace{-8pt}
\end{figure*}

\subsection{Multiple Tasks}
In practical scenarios, the goal is to train a versatile LVM capable of handling multiple tasks.
However, data imbalance poses a challenge, with some tasks having ample data while others have limited data.
Training directly with this unbalanced mixture of data from various tasks hinders overall model performance.
To address this issue, the re-sampling scheme~\cite{longtail} repeat samples from minority classes to rebalance the classes.
Following this approach, we balance the data amount of different tasks by repeating samples from tasks with limited data.
Additionally, we explore the effectiveness of using data augmentation to achieve balanced task data.
To conduct multitask experiments, we focus on three distinct vision tasks: image segmentation, human pose estimation, and image deraining.
The training data for these tasks comprises a subset of the SA-1B dataset (approximately 10\% of the entire dataset)~\cite{sam}, the complete COCO-Pose dataset~\cite{coco}, and the entire Rain13K dataset~\cite{rain13k}.
In each data setting, the model undergoes a fixed total of 35K iterations during the training process.
To evaluate the performance of our trained model, we use a withheld subset of SA-1B, along with the MPII dataset~\cite{mpii} and the Test2800 dataset~\cite{test2800}.

Table~\ref{tab:3task_aug} lists the validation results.
In comparison to training on the direct mixture of data from the three tasks, training the model on data balanced by re-sampling even results in worse performance, particularly evident in the human pose estimation and image deraining tasks with repeated samples.
By contrast, the model trained on the augmented dataset achieves better performance, notably excelling in the human pose estimation and image deraining tasks.
Notably, the use of unbalanced data yields the best quantitative results on the image segmentation task, likely because the majority of the training data belongs to this specific task.

To closely examine the impact of various dataset processing schemes in the multitask scenario, we visualize the inference results of the trained models in Figure~\ref{fig:3tasks_visualization}.
When trained with unbalanced data, the model excels in the image segmentation task.
However, for tasks of human pose estimation and image deraining, where there is less training data, its performance is subpar.
For instance, in test case 2 of the deraining task, the model seems to be confused by the abundance of segmentation samples in the training data, resulting in a segmentation output despite the presence of deraining prompts.
This emphasizes that the direct use of unbalanced data for training cannot yield excellent LVMs.
When the model is trained with task data balanced by re-sampling, it achieves even poorer performance.
In the image segmentation and deraining tasks, the model can only produce tolerable inference results in test case 1 while failing to provide informative outputs in test case 2.
Moreover, in the deraining task in both cases, it yields disordered results.
Conversely, when employing data augmentation to achieve data balance, the trained model demonstrates proficiency in all three tasks and outperforms the other two models.

\section{Knowledge Distillation}

\begin{table}[!ht]
  \renewcommand\tabcolsep{6pt}
  \caption{\textbf{KD proves beneficial in enhancing the single-task performance of LVMs.} We employ a LLaMA-1B model as the teacher to train a student model LLaMA-300M using KD. In comparison to training the student model from scratch, the introduction of KD significantly improves performance on both the image segmentation and human pose estimation tasks.}
  \label{tab:single_task_kd}
  \small
  \begin{center}
    \begin{tabular}{lcccc}
      \toprule
      \textbf{Model}    & \textbf{KD}     & \textbf{loss} $\downarrow$ & \textbf{accuracy} $\uparrow$ & \textbf{perplexity} $\downarrow$ \\
      \midrule
      \multicolumn{5}{l}{\emph{Image Segmentation (10\% of SA-1B)}}                         \\
      LLaMA-1B   & -      & 4.50              & 20.18               & \enspace90.04           \\
      LLaMA-300M & \xmark & 4.64              & 19.17               & 103.24                  \\
      LLaMA-300M & \cmark & 4.59              & 19.48               & \enspace98.72           \\
      \midrule
      \multicolumn{5}{l}{\emph{Pose Estimation (COCO-Pose)}}                                \\
      LLaMA-1B   & -      & 4.90              & 20.96               & 134.07                  \\
      LLaMA-300M & \xmark & 4.97              & 20.46               & 144.08                  \\
      LLaMA-300M & \cmark & 4.91              & 20.80               & 135.91                  \\
      \bottomrule
    \end{tabular}
  \end{center}
\end{table}

\begin{figure*}[!ht]
\begin{center}
\centerline{\includegraphics[width=0.98\linewidth]{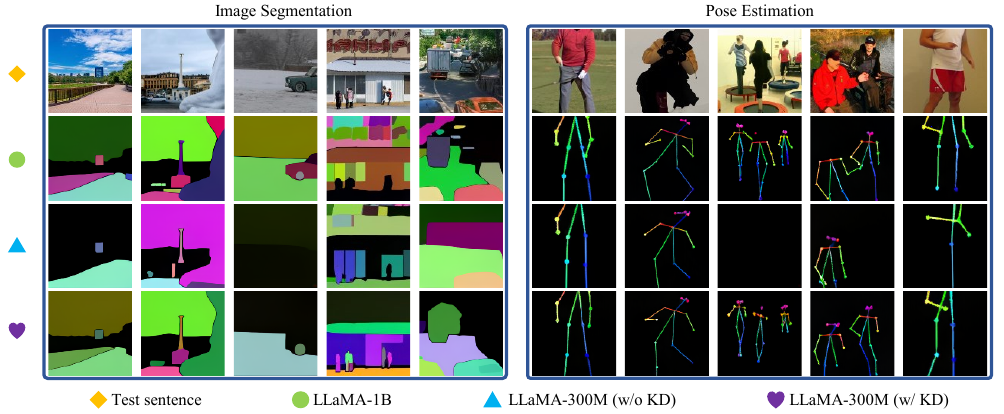}}
\vspace{-10pt}
\caption{\small{Visualization of the inference results of single-task trained models. The LLaMA-1B teacher model produces the best outcomes, and the LLaMA-300M model trained with KD exhibits greater similarity to the teacher model compared to the model trained from scratch.}}
\vspace{-20pt}
\label{fig:single_kd}
\end{center}
\end{figure*}

\begin{table*}[!ht]
  \renewcommand\tabcolsep{5pt}
  \caption{\textbf{KD also helps enhance the multi-task performance of LVMs.} We assess the effectiveness of KD with balanced data for three tasks using augmentation. KD demonstrates its usefulness by improving the performance of the student model across all tasks.}
  \vspace{-6pt}
  \label{tab:3task_kd}
  \small
  \begin{center}
    \begin{tabular}{lc|ccc|ccc|ccc}
      \toprule
      && \multicolumn{3}{c|}{\textbf{Image Segmentation}} & \multicolumn{3}{c|}{\textbf{Pose Estimation}} & \multicolumn{3}{c}{\textbf{Image Deraining}}  \\
      \textbf{Model} & \textbf{KD} & loss $\downarrow$ & accuracy $\uparrow$ & perplexity $\downarrow$ & loss $\downarrow$ & accuracy $\uparrow$ & perplexity $\downarrow$ & loss $\downarrow$ & accuracy $\uparrow$ & perplexity $\downarrow$ \\
      \midrule
      LLaMA-1B         & -           & 4.55              & 19.72               & \enspace94.75           & 4.86              & 20.77               & 129.95                  & 5.53              & 12.20               & 245.33                  \\
      LLaMA-300M       & \xmark      & 4.68              & 18.79               & 107.76                  & 4.94              & 20.24               & 140.39                  & 5.63              & 11.57               & 271.38                  \\
      LLaMA-300M       & \cmark      & 4.67              & 18.84               & 106.81                  & 4.93              & 20.32               & 139.27                  & 5.62              & 11.93               & 269.04                  \\
      \bottomrule
    \end{tabular}
    \vspace{-8pt}
  \end{center}
\end{table*}

In the quest for efficient and compact models, knowledge distillation (KD) stands out as a prevalent technique used to bolster model performance, as highlighted in~\cite{kd}. KD capitalizes on a pre-trained, larger teacher model to guide a smaller, more efficient student model to emulate the teacher's outputs. Yet, the majority of KD applications have been predominantly associated with CNN and Transformer, leaving a noticeable void in the exploration of KD for autoregerssive LVMs. Addressing this oversight, our work delves into the feasibility of applying KD in the training of LVMs, aiming to extend the benefits of KD to these streamlined models.

We begin our investigation by examining the impact of KD in single-task settings. For the teacher models, we utilize LLaMA-1B~\cite{llama}, training them on specific tasks such as image segmentation and human pose estimation. Subsequently, we train LLaMA-300M as the student models, adhering to the KD framework initially proposed by Hinton et al.~\cite{kd}. All additional experimental setting are maintained as described in the preceding section, ensuring consistency in our study.
The results of single-task KD are summarized in Table~\ref{tab:single_task_kd}.
LLaMA-1B teacher models, boasting the highest number of model parameters, demonstrates superior performance on both tasks, outperforming LLaMA-300M by a significant margin.
Nevertheless, with the incorporation of KD, the performance gap markedly diminishes in terms of validation loss, accuracy, and perplexity.
This underscores the efficacy of KD in the single-task scenario.

Correspondingly, the inference outcomes of all the models trained for both tasks are illustrated in Figure~\ref{fig:single_kd}. The visual results align with the quantitative data, demonstrating that the LLaMA-1B models, serving as the teacher models, present superior visualizations with a higher level of detail in the output images. On the other hand, the LLaMA-300M models, when trained from scratch (without the guidance of KD), yield inference results that are less detailed, and these are evidently surpassed by those models that have been trained utilizing the KD approach. 

We expand our investigation to encompass multi-task scenarios. By adhering to the experimental settings previously outlined, we employ a LLaMA-1B teacher to impart its knowledge to a LLaMA-300M student. This knowledge transfer occurs through the utilization of a combination of three datasets, all balanced via data augmentation techniques.
The validation outcomes for this multi-task scenario are summarized in Table~\ref{tab:3task_kd}. Upon comparing the performance of the distilled model with that of the model trained from scratch, we observe that the former exhibits superior results. This observation highlights the efficacy of KD in enhancing the performance of LVMs, even within the more practical and complex context of multi-task scenarios. 

In addition, we conduct a quantitative evaluation of our distilled models on a foreground segmentation task. The objective is to binary-segment a given query image into foreground and background components. The task prompt consists of three example image pairs, followed by a test (query) image. We prompt our model to generate the next 256 tokens, which are then decoded into the output image. For benchmarking, we utilize the Pascal-5i dataset established in~\cite{pascal5i}, following the methodologies of \cite{lvm,vp}. This dataset encompasses four distinct image splits, each containing between 346 and 725 images alongside corresponding segmentation masks. Each class in the dataset is represented by several image-mask pairs and is supplemented with held-out image queries for evaluation. We report our results using the mean Intersection Over Union (mIOU) metric. We examine the performance of our models in both a zero-shot setting and after finetuning. As illustrated in Table~\ref{tab:foreground_seg}, the distilled LLaMA-300M model outperforms its counterparts that were trained from scratch. Despite our model being trained on the SAM dataset, where object colors are randomized, and our use of prompt images with black backgrounds, our generated images still exhibit some randomness in color. This randomness make it challenging to separate the binary segments through post-processing. To address this, we finetuned the model using the training split of Pascal-5i, which features uniformly black backgrounds. After finetuning, we observed a significant boost in the mIOU scores.

It is important to highlight that both LVM \cite{lvm} and Visual Prompting \cite{vp} have retrained their VQGAN encoders on custom datasets, resulting in more satisfactory mIoU scores. Attaining a higher mIoU is not the primary goal of this paper; hence, we have chosen to use the encoder trained on the Laion dataset \cite{laion} directly. We believe that a better encoder and decoder would yield improved results.

\section{Ablation Study}

\begin{table}
\caption{\small{Results on Foreground Segmentation. $^\dagger$ indicates we finetune the autogressive model to generate black background.}}
\label{tab:foreground_seg}
\centering
\small
\renewcommand{\arraystretch}{1.0}
\renewcommand\tabcolsep{2pt}
\resizebox{1.0\linewidth}{!}
{
\begin{tabular}{l|rrrr}
\toprule
Model &  \multicolumn{4}{c}{Foreground Segmentation~$\uparrow$} \\
& Split 0 & Split 1 & Split 2 & Split 3  \\
\midrule
BEiT (IN-21k)~\cite{beit}    & 0.38 & 0.93 & 0.90 & 0.95 \\
VQGAN (IN-1k)~\cite{vqgan}    & 6.96 & 10.55 & 9.59 & 9.43 \\
MAE-VQGAN (IN-1k)~\cite{vp} & 2.22 & 7.07 & 5.48 & 6.28 \\
\midrule
BEiT (Figures)~\cite{vp}      & 5.38 & 3.94 & 3.20 & 3.29 \\
VQGAN (Figures)~\cite{vp}     &  12.56 & 17.51 & 14.27 & 15.06 \\
MAE-VQGAN (Figures)~\cite{vp} & 27.83 & 30.44 & 26.15 & 24.25 \\
LVM-3B (Figures)~\cite{lvm}    & 48.94 & 51.29 & 47.66 & 50.82 \\
\midrule
LLaMA-300M (Laion)         & 12.46 & 16.92 & 12.99 & 15.76 \\ 
Distilled LLaMA-300M (Laion)           & 14.72 & 17.91 & 14.55 & 17.13 \\
Distilled LLaMA-300M (Laion)$^\dagger$ & 18.58 & 21.32 & 19.90 & 21.08 \\ 
\bottomrule
\end{tabular}
}
\end{table}

\begin{figure}[!ht]
\centering
\small
\begin{tabular}{c}
\includegraphics[width=1\linewidth]{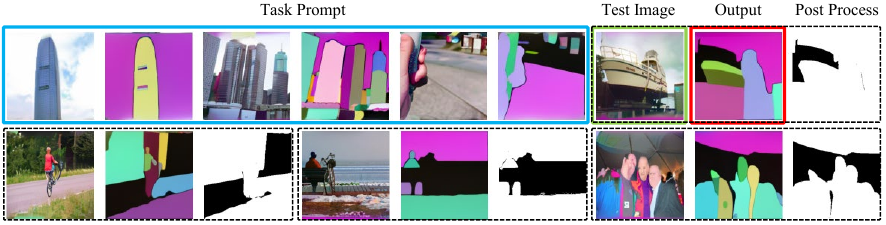} \\
(a) LLaMA-300M trained from scratch. \\
\includegraphics[width=1\linewidth]{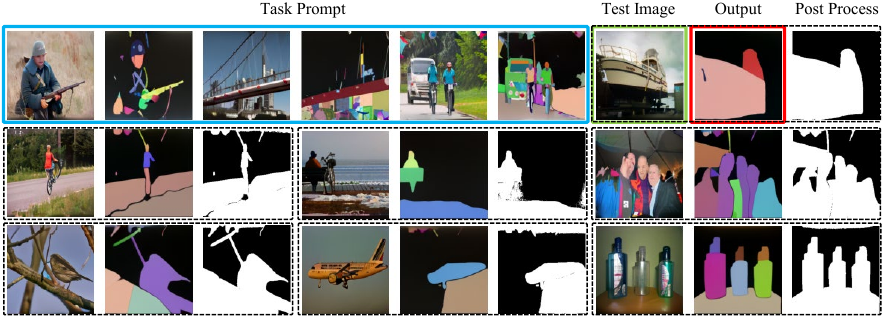} \\
(b) LLaMA-300M trained from scratch.  \\
\includegraphics[width=1\linewidth]{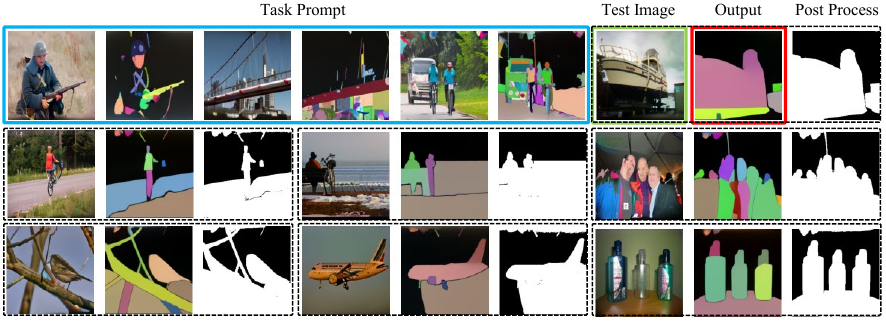} \\
(c) Distilled LLaMA-300M. \\
\includegraphics[width=1\linewidth]{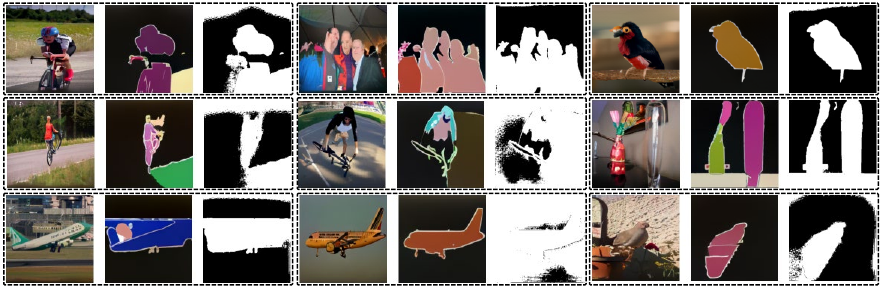} \\
(d) Finetuned distilled LLaMA-300M.\\
\end{tabular}
\vspace{-16pt}
\caption{\small{Generated output and the corresponding foreground segmentation results after our post processing. Task prompt contains a sequence of images interleaved with annotations, followed by a test image. Prompts in (a) are with a pink background, in (b) (c) (d) are with a black background.}}
\label{fig:prompt_vis}
\end{figure}

\paragraph{Influence of Different Prompts.}
In this section, we investigate the influence of different input prompts on the generated outputs in the context of foreground segmentation task. Our training dataset, SA-1B, features ground truth where each object is composed of random colors. As shown in Figure~\ref{fig:prompt_vis}(a), the segmentation ground truth for example image pairs includes a pink background. In these cases, the model tends to generate outputs with a similar pink background. Conversely, if we use a prompt with a black background, as illustrated in Figure~\ref{fig:prompt_vis}(b), the resulting image is likely to have a black background. A simple post-processing step can be applied to the output image by converting it to a grayscale image and then applying a very low threshold to produce a binary mask.

Furthermore, we finetune\footnote{Given the limited number of training images-only a few hundred—we resampled these images 100 times to construct sequential data akin to what was used during training. The entire fine-tuning process was completed in approximately 5 minutes.} our model using the training split of foreground segmentation benchmarks to focus more on the primary objects while disregarding background elements such as grass and sky. As evident from Figure~\ref{fig:prompt_vis}(d), the output of the model after fine-tuning is more concise and refined. This demonstrates the potent capability of autoregressive vision models to transfer learning effectively through a pretrain-then-finetune strategy.

\paragraph{Continual learning.}
In our preceding experiments, LVMs are trained with shuffled data to address multiple tasks.
To investigate the impact of the data shuffle operation, we train a LLaMA-300M model using an ordered concatenation of multi-task data.
Specifically, the training data follow the sequence of SA-1B, COCO-Pose, and Rain13K.
Upon completing the training for each task, we evaluate the perplexity of the model across all three tasks and present the results in Table~\ref{tab:cl}.
The results reveal that the model attains low perplexity exclusively on the most recently encountered task, exhibiting subpar performance on the other tasks, regardless of whether it has undergone training for those specific tasks.

\begin{figure}[!ht]
\begin{center}
\centerline{\includegraphics[width=0.95\linewidth]{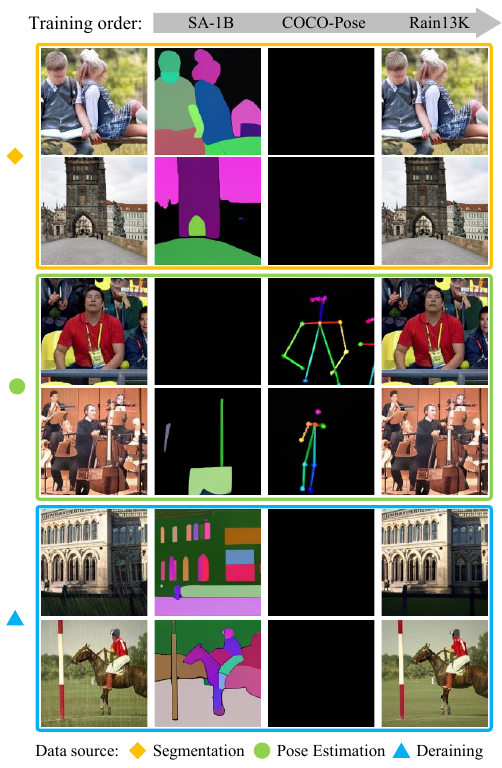}}
\vspace{-8pt}
\caption{\small{Visualization of inference results at different training stages. Task prompt is the same as in previous visualizations. The model's proficiency is focused solely on the most recently encountered training task, regardless of the provided prompt.}}
\label{fig:cl}
\end{center}
\vspace{-16pt}
\end{figure}

We present visualizations of model inference results in Figure~\ref{fig:cl}, employing the same task prompts as in previous sections.
Notably, the model tends to prioritize the inference of its last-trained task for the given inputs, seemingly disregarding the guidance provided by the task prompts.
These outcomes underscore the significance of the shuffle operation in the multi-task training of LVMs.
Essentially, LVMs face challenges related to catastrophic forgetting in continual learning scenarios.
\begin{table}[!ht]
\renewcommand{\arraystretch}{0.9}
\renewcommand\tabcolsep{4pt}
\caption{\textbf{LVM suffers from catastrophic forgetting in continual learning scenarios.} We train a LLaMA-300M without shuffling training data. The model exhibited proficient result exclusively on the task corresponding to the most recently used training data.}
\label{tab:cl}
\small
\begin{center}
\begin{tabular}{l|ccc}
  \toprule
  & \multicolumn{3}{c}{\textbf{Validation perplexity} $\downarrow$}\\
  \textbf{Data order} & Segmentation   & Pose Estimation & Deraining      \\
  \midrule
  SA-1B         & \enspace102.06 & \enspace243.09  & 1165.21        \\
  COCO-Pose     & \enspace416.21 & \enspace134.12  & 2633.08        \\
  Rain13K       & 1449.42        & 1285.80         & \enspace287.79 \\
  \bottomrule
\end{tabular}
\vspace{-16pt}
\end{center}
\end{table}

\section{Practical LLaMA-80M}

To delve deeper into the performance capabilities of efficient LVMs, we augment our exploration by training LLaMA-80M models using both data augmentation and KD.
To procure a powerful LLaMA-1B teacher, additional unlabeled image and video data are integrated into the training process.
Table~\ref{tab:llama80} lists the validation perplexity results of the trained models, revealing that the LLaMA-80M model trained with KD surpasses its counterpart trained from scratch.

We also evaluate the image understanding abilities of LLaMA-80M on ImageNet. Following the self-supervised MAE framework~\cite{mae}, we replace the VQGAN encoder with a patch embedding layer and incorporate an average pooling layer followed by a fully connected layer to perform the image classification task. Surprisingly, our LLaMA-80M achieves an impressive top-1 accuracy of 83\% on ImageNet. Although this accuracy is lower compared to some Masked Image Modeling-based methods, it outperforms training from scratch approaches. This suggests that there may be a potential for simultaneous learning of both generation and understanding tasks, indicating a promising relationship between these two aspects.

\begin{table}[!t]
\renewcommand{\arraystretch}{0.9}
\renewcommand\tabcolsep{2.5pt}
\caption{Comparison of validation perplexity between LLaMA-80M models trained with and without KD.}
\label{tab:llama80}
\small
\begin{center}
\begin{tabular}{lc|ccc}
  \toprule
  && \multicolumn{3}{c}{\textbf{Validation perplexity} $\downarrow$}\\
  \textbf{Model} & \textbf{KD} & Segmentation & Pose Estimation & Deraining \\
  \midrule
  LLaMA-1B       & -           & 88.33        & 111.95          & 230.94    \\
  LLaMA-80M      & \xmark      & 156.19       & 164.57          & 264.56    \\
  LLaMA-80M      & \cmark      & 147.78       & 159.78          & 256.65    \\
  \hline
  LLaMA-80M      & \cmark      & \multicolumn{3}{c}{ImageNet Top-1 Acc: 83.04}   \\
  \bottomrule
\end{tabular}
\vspace{-14pt}
\end{center}
\end{table}

\section{Related Works}
\textbf{Autoregressive Models.}
Autoregressive (AR) models are central to sequence prediction and have undergone significant evolution. In the realm of natural language processing (NLP), LSTM networks emerged as foundational AR models, adept at managing long-range dependencies crucial for tasks such as language modeling. The advent of the transformer architecture~\cite{transformer} marked a pivotal change in AR modeling. Its self-attention mechanisms facilitated a more efficient and parallel processing approach, which in turn, catalyzed the advancement of generative pre-training methodologies~\cite{dai2015semi,gpt3} that predict segments of text based on preceding ones. Capitalizing on their triumph in NLP, AR models have been tailored for generative image modeling tasks~\cite{uria2013rnade,van2016pixel,parmar2018image}. And many transformer-based AR vision models~\cite{igpt,lvm,vp,yu2021diverse} have achieved notable comprehension outcomes, thus validating the efficacy of AR principles for vision-related applications.

Despite the trend, much of the existing literature primarily focuses on the advantages of up-scaling model~\cite{el2024scalable,kolesnikov2019revisiting} and dataset sizes~\cite{mae,lvm}. While larger models exhibit superior performance, their heavy architectures demand extensive computational resources, thereby limiting their deployment on computation-constrained edge devices. Addressing this gap, our paper proposes a strategy aimed at crafting a more compact AR model that leverages data-efficient training techniques, offering a balance between performance and computational practicality.

\textbf{In-context based Multi-task Learning.}
Learning to perform multiple tasks simultaneously is a longstanding challenge in computer vision. Moving beyond earlier multi-task learning frameworks that depend on fixed task protocols~\cite{doersch2017multi,sener2018multi}, there are innovative attempts to integrate various vision tasks under a unified model. For example, Pix2Seq~\cite{pix2seq} pioneered this direction by treating various task outputs as elements in a discrete space. Approaches like Unified-IO~\cite{lu2022unified}, OFA~\cite{wang2022ofa}, UViM~\cite{uvim}, and Painter~\cite{painter} further streamlined this process by converting diverse inputs and outputs into sequences of tokens.

Building on this, newer methods draw inspiration from in-context learning in large language models, eliminating the task-specific paradigm altogether. These models deduce tasks directly from the input prompt, with~\cite{pathak2016context} predicting missing sections of images and~\cite{vp} merging task examples with a query image to generate outputs via inpainting. LVM~\cite{lvm} extends this concept, allowing for an expanded input sample set.

\textbf{Learning from Long-tail distribution.}
The long-tailed distribution of sample classes is a prevalent and taxing issue in computer vision~\cite{longtail}. Models trained under skewed distributions experience a marked performance drop due to the imbalance in class representation. A straightforward approach to mitigate this issue is to re-balance the training dataset such that each class is equally represented~\cite{kang2019decoupling}. Unlike previous scenarios where imbalances manifest across individual classes, our focus shifts to the disparity in the volume of data across various tasks. As a foundational strategy, we employ simple re-sampling techniques. We further explore traditional data augmentation methods and demonstrate that they not only contribute to a more equitable distribution of training samples among tasks, but also yield compelling results.

\textbf{Knowledge distillation.}
Knowledge Distillation (KD) has emerged as a prominent model compression technique over the past decade.
The core idea of KD involves leveraging the soft predictions generated by a pre-trained teacher model to guide the training of a more compact student model~\cite{kd}.
Additional loss functions have been proposed in subsequent studies to further enhance KD~\cite{dkd,dist}.
Beyond focusing solely on the final output, KD can integrate intermediate features through techniques such as pixel-level matching~\cite{fitnet,tofd}, attention alignment~\cite{komodakis2017paying}, relation matching~\cite{cc,rkd}, or contrastive learning~\cite{crd}. 
While the success of KD has been validated across various applications, its applicability to LVMs remains relatively unexplored.
Our work fills this gap and confirms the efficacy of KD within the domain of LVMs.

\section{Conclusion and Discussion}
This paper researches on developing efficient autoregression-based general-purpose vision model, oriented towards the over-reliance on colossal models and extensive balanced data. Our observation of impaired results when the model faces unevenly-distributed data across tasks underpins our data augmentation strategy to automatically enrich the limited datasets and boost the overall performance of LVMs. Furthermore,our findings demonstrate the potential of KD to bridge performance and efficiency for autoregressive LVMs.
This work serves to advance understanding into autoregressive LVMs and provide a basis for designing more efficient generalist vision models.

%\clearpage
%\section{Impact Statements}
%\textbf{Societal consequences.} This paper presents work whose goal is to advance the field of Machine Learning. There are many potential societal consequences of our work, none which we feel must be specifically highlighted here.

\textbf{Limitation and future direction.} While it is now possible to generate outputs for various vision tasks using corresponding prompts at test time, the conversion of visual results into quantifiable outputs warrants further exploration. For instance, post-processing techniques can generate foreground segmentation outputs, and in theory, color mapping can produce COCO-pose compatible outputs for keypoint detection. However, when faced with more complex tasks, it is difficult to generate quantifiable outputs from output images. Addressing this challenge represents a valuable research direction. Finding effective ways to derive quantifiable outputs for different tasks from VQGAN-decoded images is an area that requires further study.

% \nocite{xxx}

\small
{
\bibliography{egbib}

\begin{thebibliography}{47}
\providecommand{\natexlab}[1]{#1}
\providecommand{\url}[1]{\texttt{#1}}
\expandafter\ifx\csname urlstyle\endcsname\relax
  \providecommand{\doi}[1]{doi: #1}\else
  \providecommand{\doi}{doi: \begingroup \urlstyle{rm}\Url}\fi

\bibitem[Achiam et~al.(2023)Achiam, Adler, Agarwal, Ahmad, Akkaya, Aleman,
  Almeida, Altenschmidt, Altman, Anadkat, et~al.]{gpt4}
Achiam, J., Adler, S., Agarwal, S., Ahmad, L., Akkaya, I., Aleman, F.~L.,
  Almeida, D., Altenschmidt, J., Altman, S., Anadkat, S., et~al.
\newblock Gpt-4 technical report.
\newblock \emph{arXiv preprint arXiv:2303.08774}, 2023.

\bibitem[Andriluka et~al.(2014)Andriluka, Pishchulin, Gehler, and
  Schiele]{mpii}
Andriluka, M., Pishchulin, L., Gehler, P., and Schiele, B.
\newblock 2d human pose estimation: New benchmark and state of the art
  analysis.
\newblock In \emph{Proceedings of the IEEE Conference on computer Vision and
  Pattern Recognition}, pp.\  3686--3693, 2014.

\bibitem[Bai et~al.(2023)Bai, Geng, Mangalam, Bar, Yuille, Darrell, Malik, and
  Efros]{lvm}
Bai, Y., Geng, X., Mangalam, K., Bar, A., Yuille, A., Darrell, T., Malik, J.,
  and Efros, A.~A.
\newblock Sequential modeling enables scalable learning for large vision
  models.
\newblock \emph{arXiv preprint arXiv:2312.00785}, 2023.

\bibitem[Bao et~al.(2021)Bao, Dong, Piao, and Wei]{beit}
Bao, H., Dong, L., Piao, S., and Wei, F.
\newblock Beit: Bert pre-training of image transformers.
\newblock \emph{arXiv preprint arXiv:2106.08254}, 2021.

\bibitem[Bar et~al.(2022)Bar, Gandelsman, Darrell, Globerson, and Efros]{vp}
Bar, A., Gandelsman, Y., Darrell, T., Globerson, A., and Efros, A.
\newblock Visual prompting via image inpainting.
\newblock \emph{Advances in Neural Information Processing Systems}, 2022.

\bibitem[Brown et~al.(2020)Brown, Mann, Ryder, Subbiah, Kaplan, Dhariwal,
  Neelakantan, Shyam, Sastry, Askell, et~al.]{gpt3}
Brown, T., Mann, B., Ryder, N., Subbiah, M., Kaplan, J.~D., Dhariwal, P.,
  Neelakantan, A., Shyam, P., Sastry, G., Askell, A., et~al.
\newblock Language models are few-shot learners.
\newblock \emph{Advances in neural information processing systems}, 2020.

\bibitem[Chang et~al.(2023)Chang, Zhang, Barber, Maschinot, Lezama, Jiang,
  Yang, Murphy, Freeman, Rubinstein, et~al.]{muse}
Chang, H., Zhang, H., Barber, J., Maschinot, A., Lezama, J., Jiang, L., Yang,
  M.-H., Murphy, K., Freeman, W.~T., Rubinstein, M., et~al.
\newblock Muse: Text-to-image generation via masked generative transformers.
\newblock \emph{arXiv preprint arXiv:2301.00704}, 2023.

\bibitem[Chen et~al.(2020)Chen, Radford, Child, Wu, Jun, Luan, and
  Sutskever]{igpt}
Chen, M., Radford, A., Child, R., Wu, J., Jun, H., Luan, D., and Sutskever, I.
\newblock Generative pretraining from pixels.
\newblock In \emph{International conference on machine learning}, 2020.

\bibitem[Chen et~al.(2021)Chen, Saxena, Li, Fleet, and Hinton]{pix2seq}
Chen, T., Saxena, S., Li, L., Fleet, D.~J., and Hinton, G.
\newblock Pix2seq: A language modeling framework for object detection.
\newblock \emph{arXiv preprint arXiv:2109.10852}, 2021.

\bibitem[Chowdhery et~al.(2023)Chowdhery, Narang, Devlin, Bosma, Mishra,
  Roberts, Barham, Chung, Sutton, Gehrmann, et~al.]{palm}
Chowdhery, A., Narang, S., Devlin, J., Bosma, M., Mishra, G., Roberts, A.,
  Barham, P., Chung, H.~W., Sutton, C., Gehrmann, S., et~al.
\newblock Palm: Scaling language modeling with pathways.
\newblock \emph{Journal of Machine Learning Research}, 2023.

\bibitem[Dai \& Le(2015)Dai and Le]{dai2015semi}
Dai, A.~M. and Le, Q.~V.
\newblock Semi-supervised sequence learning.
\newblock \emph{Advances in neural information processing systems}, 2015.

\bibitem[Doersch \& Zisserman(2017)Doersch and Zisserman]{doersch2017multi}
Doersch, C. and Zisserman, A.
\newblock Multi-task self-supervised visual learning.
\newblock In \emph{Proceedings of the IEEE international conference on computer
  vision}, 2017.

\bibitem[El-Nouby et~al.(2024)El-Nouby, Klein, Zhai, Bautista, Toshev, Shankar,
  Susskind, and Joulin]{el2024scalable}
El-Nouby, A., Klein, M., Zhai, S., Bautista, M.~A., Toshev, A., Shankar, V.,
  Susskind, J.~M., and Joulin, A.
\newblock Scalable pre-training of large autoregressive image models.
\newblock \emph{arXiv preprint arXiv:2401.08541}, 2024.

\bibitem[Esser et~al.(2021)Esser, Rombach, and Ommer]{vqgan}
Esser, P., Rombach, R., and Ommer, B.
\newblock Taming transformers for high-resolution image synthesis.
\newblock In \emph{Proceedings of the IEEE/CVF conference on computer vision
  and pattern recognition}, 2021.

\bibitem[Fu et~al.(2017)Fu, Huang, Zeng, Huang, Ding, and Paisley]{test2800}
Fu, X., Huang, J., Zeng, D., Huang, Y., Ding, X., and Paisley, J.
\newblock Removing rain from single images via a deep detail network.
\newblock In \emph{Proceedings of the IEEE conference on computer vision and
  pattern recognition}, pp.\  3855--3863, 2017.

\bibitem[He et~al.(2022)He, Chen, Xie, Li, Doll{\'a}r, and Girshick]{mae}
He, K., Chen, X., Xie, S., Li, Y., Doll{\'a}r, P., and Girshick, R.
\newblock Masked autoencoders are scalable vision learners.
\newblock In \emph{Proceedings of the IEEE/CVF conference on computer vision
  and pattern recognition}, 2022.

\bibitem[Hinton et~al.(2015)Hinton, Vinyals, and Dean]{kd}
Hinton, G.~E., Vinyals, O., and Dean, J.
\newblock Distilling the knowledge in a neural network.
\newblock \emph{arXiv preprint arXiv: 1503.02531}, 2015.

\bibitem[Huang et~al.(2022)Huang, You, Wang, Qian, and Xu]{dist}
Huang, T., You, S., Wang, F., Qian, C., and Xu, C.
\newblock Knowledge distillation from a stronger teacher.
\newblock In \emph{Advances in Neural Information Processing Systems}, 2022.

\bibitem[Jiang et~al.(2020)Jiang, Wang, Yi, Chen, Huang, Luo, Ma, and
  Jiang]{rain13k}
Jiang, K., Wang, Z., Yi, P., Chen, C., Huang, B., Luo, Y., Ma, J., and Jiang,
  J.
\newblock Multi-scale progressive fusion network for single image deraining.
\newblock In \emph{Proceedings of the IEEE/CVF conference on computer vision
  and pattern recognition}, pp.\  8346--8355, 2020.

\bibitem[Kang et~al.(2019)Kang, Xie, Rohrbach, Yan, Gordo, Feng, and
  Kalantidis]{kang2019decoupling}
Kang, B., Xie, S., Rohrbach, M., Yan, Z., Gordo, A., Feng, J., and Kalantidis,
  Y.
\newblock Decoupling representation and classifier for long-tailed recognition.
\newblock \emph{arXiv preprint arXiv:1910.09217}, 2019.

\bibitem[Kirillov et~al.(2023)Kirillov, Mintun, Ravi, Mao, Rolland, Gustafson,
  Xiao, Whitehead, Berg, Lo, et~al.]{sam}
Kirillov, A., Mintun, E., Ravi, N., Mao, H., Rolland, C., Gustafson, L., Xiao,
  T., Whitehead, S., Berg, A.~C., Lo, W.-Y., et~al.
\newblock Segment anything.
\newblock \emph{arXiv preprint arXiv:2304.02643}, 2023.

\bibitem[Kolesnikov et~al.(2019)Kolesnikov, Zhai, and
  Beyer]{kolesnikov2019revisiting}
Kolesnikov, A., Zhai, X., and Beyer, L.
\newblock Revisiting self-supervised visual representation learning.
\newblock In \emph{Proceedings of the IEEE/CVF conference on computer vision
  and pattern recognition}, 2019.

\bibitem[Kolesnikov et~al.(2022)Kolesnikov, Susano~Pinto, Beyer, Zhai, Harmsen,
  and Houlsby]{uvim}
Kolesnikov, A., Susano~Pinto, A., Beyer, L., Zhai, X., Harmsen, J., and
  Houlsby, N.
\newblock Uvim: A unified modeling approach for vision with learned guiding
  codes.
\newblock \emph{Advances in Neural Information Processing Systems}, 2022.

\bibitem[Komodakis \& Zagoruyko(2017)Komodakis and
  Zagoruyko]{komodakis2017paying}
Komodakis, N. and Zagoruyko, S.
\newblock Paying more attention to attention: improving the performance of
  convolutional neural networks via attention transfer.
\newblock In \emph{International Conference on Learning Representations}, 2017.

\bibitem[Lin et~al.(2014)Lin, Maire, Belongie, Hays, Perona, Ramanan,
  Doll{\'a}r, and Zitnick]{coco}
Lin, T.-Y., Maire, M., Belongie, S., Hays, J., Perona, P., Ramanan, D.,
  Doll{\'a}r, P., and Zitnick, C.~L.
\newblock Microsoft coco: Common objects in context.
\newblock In \emph{Computer Vision--ECCV 2014: 13th European Conference,
  Zurich, Switzerland, September 6-12, 2014, Proceedings, Part V 13}, 2014.

\bibitem[Lu et~al.(2022)Lu, Clark, Zellers, Mottaghi, and
  Kembhavi]{lu2022unified}
Lu, J., Clark, C., Zellers, R., Mottaghi, R., and Kembhavi, A.
\newblock Unified-io: A unified model for vision, language, and multi-modal
  tasks.
\newblock \emph{arXiv preprint arXiv:2206.08916}, 2022.

\bibitem[Park et~al.(2019)Park, Kim, Lu, and Cho]{rkd}
Park, W., Kim, D., Lu, Y., and Cho, M.
\newblock Relational knowledge distillation.
\newblock In \emph{IEEE/CVF Conference on Computer Vision and Pattern
  Recognition}, 2019.

\bibitem[Parmar et~al.(2018)Parmar, Vaswani, Uszkoreit, Kaiser, Shazeer, Ku,
  and Tran]{parmar2018image}
Parmar, N., Vaswani, A., Uszkoreit, J., Kaiser, L., Shazeer, N., Ku, A., and
  Tran, D.
\newblock Image transformer.
\newblock In \emph{International conference on machine learning}, 2018.

\bibitem[Pathak et~al.(2016)Pathak, Krahenbuhl, Donahue, Darrell, and
  Efros]{pathak2016context}
Pathak, D., Krahenbuhl, P., Donahue, J., Darrell, T., and Efros, A.~A.
\newblock Context encoders: Feature learning by inpainting.
\newblock In \emph{Proceedings of the IEEE conference on computer vision and
  pattern recognition}, 2016.

\bibitem[Peng et~al.(2019)Peng, Jin, Liu, Li, Wu, Liu, Zhou, and Zhang]{cc}
Peng, B., Jin, X., Liu, J., Li, D., Wu, Y., Liu, Y., Zhou, S., and Zhang, Z.
\newblock Correlation congruence for knowledge distillation.
\newblock In \emph{International Conference on Computer Vision}, 2019.

\bibitem[Romero et~al.(2014)Romero, Ballas, Kahou, Chassang, Gatta, and
  Bengio]{fitnet}
Romero, A., Ballas, N., Kahou, S.~E., Chassang, A., Gatta, C., and Bengio, Y.
\newblock Fitnets: Hints for thin deep nets.
\newblock \emph{arXiv preprint arXiv:1412.6550}, 2014.

\bibitem[Schuhmann et~al.(2022)Schuhmann, Beaumont, Vencu, Gordon, Wightman,
  Cherti, Coombes, Katta, Mullis, Wortsman, et~al.]{laion}
Schuhmann, C., Beaumont, R., Vencu, R., Gordon, C., Wightman, R., Cherti, M.,
  Coombes, T., Katta, A., Mullis, C., Wortsman, M., et~al.
\newblock Laion-5b: An open large-scale dataset for training next generation
  image-text models.
\newblock \emph{Advances in Neural Information Processing Systems}, 2022.

\bibitem[Sener \& Koltun(2018)Sener and Koltun]{sener2018multi}
Sener, O. and Koltun, V.
\newblock Multi-task learning as multi-objective optimization.
\newblock \emph{Advances in neural information processing systems}, 2018.

\bibitem[Shaban et~al.(2017)Shaban, Bansal, Liu, Essa, and Boots]{pascal5i}
Shaban, A., Bansal, S., Liu, Z., Essa, I., and Boots, B.
\newblock One-shot learning for semantic segmentation.
\newblock \emph{arXiv preprint arXiv:1709.03410}, 2017.

\bibitem[Team(2023)]{internlm}
Team, I.
\newblock Internlm: A multilingual language model with progressively enhanced
  capabilities, 2023.

\bibitem[Tian et~al.(2020)Tian, Krishnan, and Isola]{crd}
Tian, Y., Krishnan, D., and Isola, P.
\newblock Contrastive representation distillation.
\newblock In \emph{IEEE/CVF International Conference on Learning
  Representations}, 2020.

\bibitem[Touvron et~al.(2023{\natexlab{a}})Touvron, Lavril, Izacard, Martinet,
  Lachaux, Lacroix, Rozi{\`e}re, Goyal, Hambro, Azhar, et~al.]{llama}
Touvron, H., Lavril, T., Izacard, G., Martinet, X., Lachaux, M.-A., Lacroix,
  T., Rozi{\`e}re, B., Goyal, N., Hambro, E., Azhar, F., et~al.
\newblock Llama: Open and efficient foundation language models.
\newblock \emph{arXiv preprint arXiv:2302.13971}, 2023{\natexlab{a}}.

\bibitem[Touvron et~al.(2023{\natexlab{b}})Touvron, Martin, Stone, Albert,
  Almahairi, Babaei, Bashlykov, Batra, Bhargava, Bhosale, et~al.]{llama2}
Touvron, H., Martin, L., Stone, K., Albert, P., Almahairi, A., Babaei, Y.,
  Bashlykov, N., Batra, S., Bhargava, P., Bhosale, S., et~al.
\newblock Llama 2: Open foundation and fine-tuned chat models.
\newblock \emph{arXiv preprint arXiv:2307.09288}, 2023{\natexlab{b}}.

\bibitem[Uria et~al.(2013)Uria, Murray, and Larochelle]{uria2013rnade}
Uria, B., Murray, I., and Larochelle, H.
\newblock Rnade: The real-valued neural autoregressive density-estimator.
\newblock \emph{Advances in Neural Information Processing Systems}, 2013.

\bibitem[Van Den~Oord et~al.(2016)Van Den~Oord, Kalchbrenner, and
  Kavukcuoglu]{van2016pixel}
Van Den~Oord, A., Kalchbrenner, N., and Kavukcuoglu, K.
\newblock Pixel recurrent neural networks.
\newblock In \emph{International conference on machine learning}, 2016.

\bibitem[Vaswani et~al.(2017)Vaswani, Shazeer, Parmar, Uszkoreit, Jones, Gomez,
  Kaiser, and Polosukhin]{transformer}
Vaswani, A., Shazeer, N., Parmar, N., Uszkoreit, J., Jones, L., Gomez, A.~N.,
  Kaiser, {\L}., and Polosukhin, I.
\newblock Attention is all you need.
\newblock \emph{Advances in neural information processing systems}, 2017.

\bibitem[Wang et~al.(2022)Wang, Yang, Men, Lin, Bai, Li, Ma, Zhou, Zhou, and
  Yang]{wang2022ofa}
Wang, P., Yang, A., Men, R., Lin, J., Bai, S., Li, Z., Ma, J., Zhou, C., Zhou,
  J., and Yang, H.
\newblock Ofa: Unifying architectures, tasks, and modalities through a simple
  sequence-to-sequence learning framework.
\newblock In \emph{International Conference on Machine Learning}, 2022.

\bibitem[Wang et~al.(2023)Wang, Wang, Cao, Shen, and Huang]{painter}
Wang, X., Wang, W., Cao, Y., Shen, C., and Huang, T.
\newblock Images speak in images: A generalist painter for in-context visual
  learning.
\newblock In \emph{Proceedings of the IEEE/CVF Conference on Computer Vision
  and Pattern Recognition}, 2023.

\bibitem[Yu et~al.(2021)Yu, Zhan, Wu, Pan, Cui, Lu, Ma, Xie, and
  Miao]{yu2021diverse}
Yu, Y., Zhan, F., Wu, R., Pan, J., Cui, K., Lu, S., Ma, F., Xie, X., and Miao,
  C.
\newblock Diverse image inpainting with bidirectional and autoregressive
  transformers.
\newblock In \emph{Proceedings of the 29th ACM International Conference on
  Multimedia}, 2021.

\bibitem[Zhang et~al.(2020)Zhang, Shi, Shi, Ma, and Bao]{tofd}
Zhang, L., Shi, Y., Shi, Z., Ma, K., and Bao, C.
\newblock Task-oriented feature distillation.
\newblock In \emph{Advances in Neural Information Processing Systems}, 2020.

\bibitem[Zhang et~al.(2023)Zhang, Kang, Hooi, Yan, and Feng]{longtail}
Zhang, Y., Kang, B., Hooi, B., Yan, S., and Feng, J.
\newblock Deep long-tailed learning: A survey.
\newblock \emph{IEEE Transactions on Pattern Analysis and Machine
  Intelligence}, 2023.

\bibitem[Zhao et~al.(2022)Zhao, Cui, Song, Qiu, and Liang]{dkd}
Zhao, B., Cui, Q., Song, R., Qiu, Y., and Liang, J.
\newblock Decoupled knowledge distillation.
\newblock In \emph{IEEE/CVF Conference on Computer Vision and Pattern
  Recognition}, 2022.

\end{thebibliography}
\bibliographystyle{icml2024}
}
%%%%%%%%%%%%%%%%%%%%%%%%%%%%%%%%%%%%%%%%%%%%%%%%%%%%%%%%%%%%%%%%%%%%%%%%%%%%%%%
%%%%%%%%%%%%%%%%%%%%%%%%%%%%%%%%%%%%%%%%%%%%%%%%%%%%%%%%%%%%%%%%%%%%%%%%%%%%%%%
% APPENDIX
%%%%%%%%%%%%%%%%%%%%%%%%%%%%%%%%%%%%%%%%%%%%%%%%%%%%%%%%%%%%%%%%%%%%%%%%%%%%%%%
%%%%%%%%%%%%%%%%%%%%%%%%%%%%%%%%%%%%%%%%%%%%%%%%%%%%%%%%%%%%%%%%%%%%%%%%%%%%%%%
\newpage
\appendix
\onecolumn

\section{Training Configurations}

\subsection{Model architecture}

In our main paper, we employ three LVMs with different parameters.
These models are structured based on LLaMA~\cite{llama}. The detailed architecture configurations are outlined in Table~\ref{tab:supp:model_setting}.

\begin{table}[!ht]
  \caption{Detailed configurations of used models.}
  \label{tab:supp:model_setting}
  \small
  \begin{center}
    \begin{tabular}{l|cccc}
      \toprule
      \textbf{Model} &  \textbf{Hidden dim.} & \textbf{MLP dim.} & \textbf{\#heads} & \textbf{\#layers} \\
      \midrule
      LLaMA-1B           & 2048                 & 5504              & 16               & 22                \\
      LLaMA-300M         & 1024                 & 2688              & 8                & 22                \\
      LLaMA-80M             & 768                  & 3072              & 12               & 12                \\
      \bottomrule
    \end{tabular}
  \end{center}
\end{table}

\subsection{Training details}

Our training strategy adheres to the implementation of LVM~\cite{lvm}, with slight adjustments made for efficient training with 8-16 A100 GPUs. Our models are trained based on the InternLM framework~\cite{internlm}.
The optimization details are summarized in Table~\ref{tab:supp:optim_setting}.
Table~\ref{tab:supp:cost} presents the time consumption and memory requirements for training each model.

\begin{table}[!ht]
  \caption{Detailed configurations for training efficient LVMs. We attain a consistent equivalent batch size across different models by adjusting the number of employed GPUs, mini-batch size, and gradient accumulation steps.}
  \label{tab:supp:optim_setting}
  \small
  \begin{center}
    \begin{tabular}{l|l}
      \toprule
      \textbf{Config}                & \textbf{Value}                 \\
      \midrule
      optimizer                      & AdamW                          \\
      learning rate                  & 1.5e-4                         \\
      weight decay                   & 0.1                            \\
      optimizer momentum             & $\beta_1$, $\beta_2$=0.9, 0.95 \\
      equivalent batch size (tokens) & 262144                         \\
      learning rate schedule         & cosine                         \\
      warmup steps                   & \#total steps * 0.0056         \\
      final learning rate            & 1.5e-5                         \\
      context length                 & 2048                           \\
      data augmentation              & RandomResizedCrop              \\
      \bottomrule
    \end{tabular}
  \end{center}
\end{table}

\begin{table}[!ht]
  \caption{Training time and memory requirements of each LVM. The equivalent batch size is calculated as the product of the number of utilized GPUs, mini-batch size, and gradient accumulation steps, resulting in a uniform value of 262144 across various models.}
  \label{tab:supp:cost}
  \small
  \begin{center}
    \begin{tabular}{ll|ccccc}
      \toprule
      \textbf{Model} & \textbf{Teacher} & \textbf{\#GPUs} & \textbf{Mini batch (tokens)} & \textbf{\#gradient accum.} & \textbf{Time (hours)} & \textbf{Memory (GB)} \\
      \midrule
      LLaMA-1B         & -                & 16              & 32768                        & 4                          & 324                   & 70                   \\
      LLaMA-300M       & -                & 16              & 65536                        & 2                          & 126                   & 70                   \\
      LLaMA-300M       & LLaMA-1B           & 16              & 65536                        & 2                          & 235                   & 80                   \\
      LLaMA-80M        & -                & 8               & 131072                       & 2                          & 65                    & 69                   \\
      LLaMA-80M        & LLaMA-1B           & 8               & 65536                        & 4                          & 82                    & 49                   \\
      \bottomrule
    \end{tabular}
  \end{center}
\end{table}

\subsection{Dataset}
Our full training involves diverse datasets, including Rain13K, SA-1B, COCO-pose, HDvila-100m, and LAION datasets.
To ensure data balance, data augmentation is applied to extend the Rain13K and COCO-pose datasets.

\textbf{Rain13K (16.96\%; 14.02 billion tokens).}
Rain13K serves as the most commonly utilized training dataset for rain removal, consisting of five distinct rain removal datasets. The original dataset comprises 13,712 clean-rain image pairs. We filtered the data and generated a subset for training.

\textbf{SA-1B (12.92\%; 10.68 billion tokens).}
SA-1B is a large-scale multimodal dataset designed for training general-purpose object segmentation models. It comprises 1.1 billion high-resolution, diverse, and privacy-protected images, along with corresponding high-quality segmentation masks. We selected a portion of the data for training.

\textbf{COCO-Pose (35.05\%; 28.97 billion tokens).}
COCO-Pose is a dataset for human pose detection. The original version is a subset of COCO dataset and has 250K images, with annotations for 17 human keypoints, such as eyes, hands, legs, foots, etc. COCO keypoints dataset can be used to train and evaluate various human pose detection models.

\textbf{HDvila-100m (1.25\%; 1.03 billion tokens).}
HDvila-100m is a large-scale video-language multimodal dataset containing 100 million high-resolution, diverse, and video clips and 100 million automatically generated text descriptions. The dataset covers a wide range of topics and includes both video segments and corresponding text captions. The video segments are 10 seconds long, the text captions are produced by an automatic speech recognition system. We exclusively utilized a subset of the video data to endow our model with the capability for continuous inference.

\textbf{LAION-400M (33.83\%; 27.96 billion tokens).}
LAION-400M is a large-scale multimodal dataset containing 400 million English image-text pairs. It covers a wide range of topics and includes both text descriptions and corresponding images. We adopt a subset of this dataset in our experiments.

\section{Continual Learning.}

\subsection{Additional results}

In our ablation study on data shuffling, we observe catastrophic forgetting in LVMs within an ordered data setting.
To further investigate, we assess the impact of rescaling the learning rate at the beginning of each task, a configuration more aligned with the standard continual learning setting. The summarized training configurations are presented in Table~\ref{tab:supp:cl_setting}.

\begin{figure}[!ht]
  % \vskip 0.2in
  \begin{center}
    \centerline{\includegraphics[width=\linewidth]{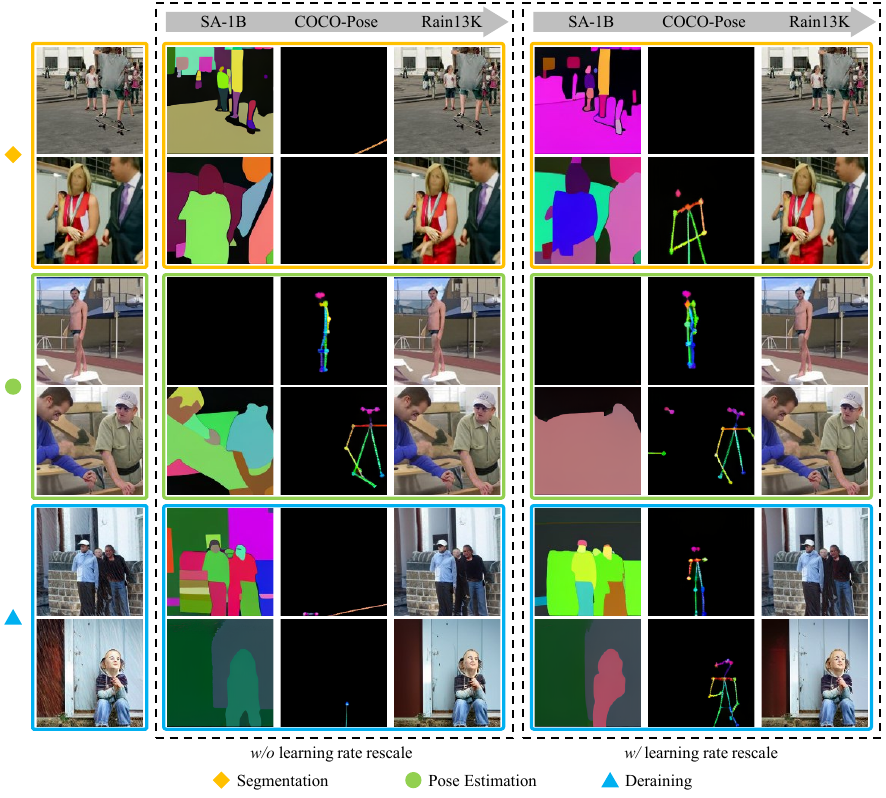}}
    \caption{Visualization of inference results of model at various training stages. Catastrophic forgetting is evident in scenarios with and without learning rate rescaling.}
    \label{fig:supp:cl}
  \end{center}
  \vskip -0.2in
\end{figure}

Table~\ref{tab:supp:cl} presents quantitative results for both scenarios—without and with learning rate rescaling—while Figure~\ref{fig:supp:cl} showcases the corresponding visualizations.
In both settings, the results indicate the presence of catastrophic forgetting in LVMs during continual learning scenarios.
Moreover, the adoption of learning rate rescaling results in improved performance on the last-trained task but exacerbates forgetting of other tasks.

\begin{table}[!ht]
  \renewcommand\tabcolsep{4pt}
  \caption{Learning rate configuration for the continual learning setting. The setup labeled as ``w/o learning rate rescaling'' corresponds to the experiment detailed in the main paper.}
  \label{tab:supp:cl_setting}
  \small
  \begin{center}
    \begin{tabular}{l|ccc}
      \toprule
      & \multicolumn{3}{c}{\textbf{Learning rate scale at task \#}} \\
      \textbf{Learning rate configuration} & Task 1 (SA-1B)     & Task 2 (COCO-Pose) & Task 3 (Rain13K)   \\
      \midrule
      w/o learning rate rescaling            & 1.5e-4$\sim$8e-5   & 8e-5$\sim$5e-5     & 5e-5$\sim$1.5e-5   \\
      w/ learning rate rescaling             & 1.5e-4$\sim$1.5e-5 & 1.5e-4$\sim$1.5e-5 & 1.5e-4$\sim$1.5e-5 \\
      \bottomrule
    \end{tabular}
  \end{center}
\end{table}

\begin{table}[!ht]
  \renewcommand\tabcolsep{4pt}
  \caption{Performance of LVMs in continual learning scenarios. We train LLaMA-300M models without shuffling the training data. Two distinct learning rate scheduling schemes were employed. Under each scheme, the LVM exhibits signs of catastrophic forgetting.}
  \label{tab:supp:cl}
  \small
  \begin{center}
    \begin{tabular}{l|ccc|ccc|ccc}
      \toprule
      & \multicolumn{3}{c|}{\textbf{Loss} $\downarrow$}& \multicolumn{3}{c|}{\textbf{Accuracy} $\uparrow$}& \multicolumn{3}{c}{\textbf{perplexity} $\downarrow$}\\
      \textbf{Data} & Segmentation & Pose Est. & Deraining & Segmentation & Pose Est.    & Deraining    & Segmentation   & Pose Est.      & Deraining      \\
      \midrule
      \multicolumn{10}{l}{\emph{w/o learning rate rescaling}}                                                                                           \\
      SA-1B         & 4.62         & 5.49      & 7.07      & 19.17        & 16.54        & \enspace1.42 & \enspace102.06 & \enspace243.09 & 1165.21        \\
      COCO-Pose     & 6.02         & 4.89      & 7.88      & 10.29        & 20.69        & \enspace0.57 & \enspace416.21 & \enspace134.12 & 2633.08        \\
      Rain13K       & 7.28         & 7.16      & 5.70      & \enspace4.45 & \enspace5.99 & 11.79        & 1449.42        & 1285.80        & \enspace287.79 \\
      \midrule
      \multicolumn{10}{l}{\emph{w/ learning rate rescaling}}                                                                                            \\
      SA-1B         & 4.62         & 5.48      & 7.11      & 19.29        & 16.65        & \enspace1.34 & \enspace101.52 & \enspace241.59 & 1213.27        \\
      COCO-Pose     & 6.22         & 4.88      & 7.98      & \enspace9.23 & 20.84        & \enspace0.54 & \enspace510.03 & \enspace132.78 & 2880.20        \\
      Rain13K       & 7.98         & 7.78      & 5.67      & \enspace2.88 & \enspace3.92 & 11.91        & 2925.92        & 2403.03        & \enspace281.81 \\
      \bottomrule
    \end{tabular}
  \end{center}
\end{table}

\subsection{Offline training}

\begin{figure}[!ht]
  % \vskip 0.2in
  \begin{center}
    \centerline{\includegraphics[width=\linewidth]{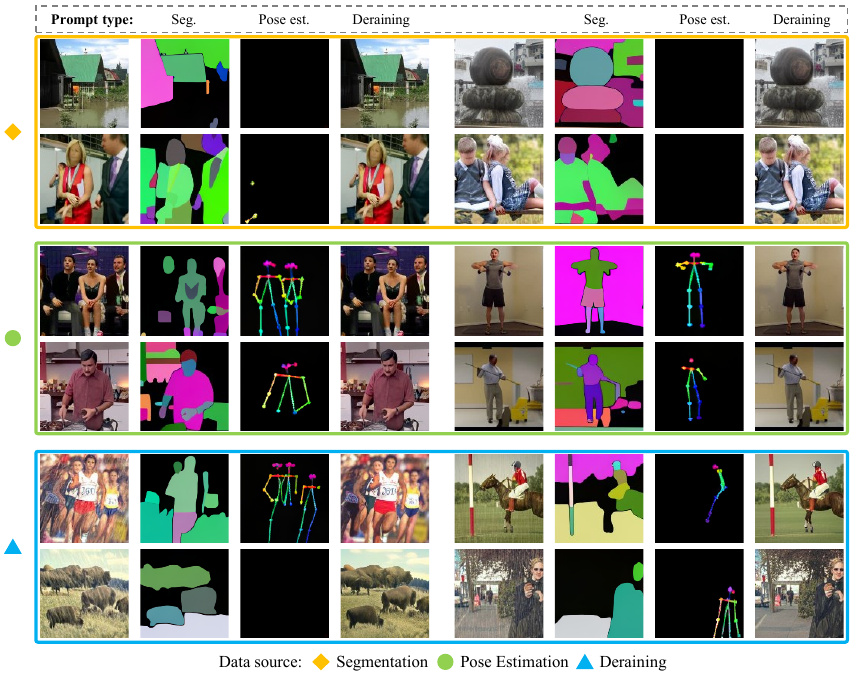}}
    \caption{Visualization of inference results of LLaMA-1B with different prompts on the same inputs.}
    \label{fig:supp:prompt}
  \end{center}
  \vskip -0.2in
\end{figure}

In Figure~\ref{fig:supp:prompt}, we compare the inference results of the offline trained LLaMA-1B model on the entire dataset using different prompts but the same inputs.
In contrast to the continual learning scenario, the model trained with shuffled data demonstrates successful recognition of the given prompts.
These results underscore the ability of the model to adeptly handle multi-task scenarios with distinct prompts.

\section{Full Results of LLaMA-80M.}

We train an LLaMA-80M model employing KD and present its performance in terms of perplexity in the main paper.

\begin{figure}[!ht]
  % \vskip 0.2in
  \begin{center}
    \centerline{\includegraphics[width=\linewidth]{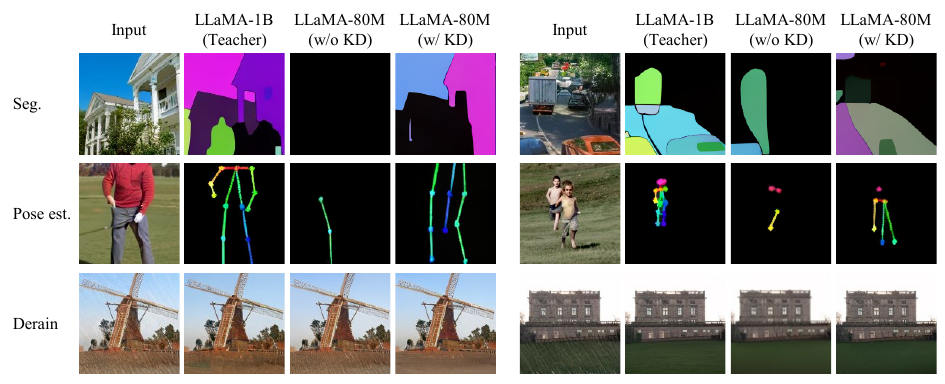}}
    \caption{Visualization of inference results from the efficient LLaMA-80M model.}
    \label{fig:llama80}
  \end{center}
  \vskip -0.2in
\end{figure}

Figure~\ref{fig:llama80} provides additional visualization of the generated images.
The efficient 80M model exhibits significant potential in handling prompted image autoregressive tasks, and the performance is further enhanced when KD is applied.
We posit that by incorporating more high-quality data and extending the training schedule, these efficient models can achieve practical success in the future.

\begin{table}[!ht]
  \renewcommand\tabcolsep{3pt}
  \caption{Comparison of validation performance between LLaMA-80M models trained with and without KD.}
  \label{tab:supp:llama80}
  \small
  \begin{center}
    \begin{tabular}{lc|ccc|ccc|ccc}
      \toprule
      && \multicolumn{3}{c|}{\textbf{Loss} $\downarrow$}& \multicolumn{3}{c|}{\textbf{Accuracy} $\uparrow$}& \multicolumn{3}{c}{\textbf{perplexity} $\downarrow$}\\
      \textbf{Data} & \textbf{KD} & Segmentation & Pose Est. & Deraining & Segmentation & Pose Est. & Deraining & Segmentation & Pose Est. & Deraining \\
      \midrule
      LLaMA-1B      & -           & 4.48         & 4.71      & 5.47      & 19.89        & 21.45     & 12.46     & 88.33        & 111.95    & 230.94    \\
      LLaMA-80M     & \xmark      & 5.04         & 5.10      & 5.60      & 16.08        & 19.03     & 12.05     & 156.19       & 164.57    & 264.56    \\
      LLaMA-80M     & \cmark      & 4.99         & 5.07      & 5.57      & 16.30        & 19.04     & 12.13     & 147.78       & 159.78    & 256.65    \\
      \bottomrule
    \end{tabular}
  \end{center}
\end{table}

\section{Single-image Inpainting}

\begin{figure}[!ht]
  % \vskip 0.2in
  \begin{center}
    \centerline{\includegraphics[width=\linewidth]{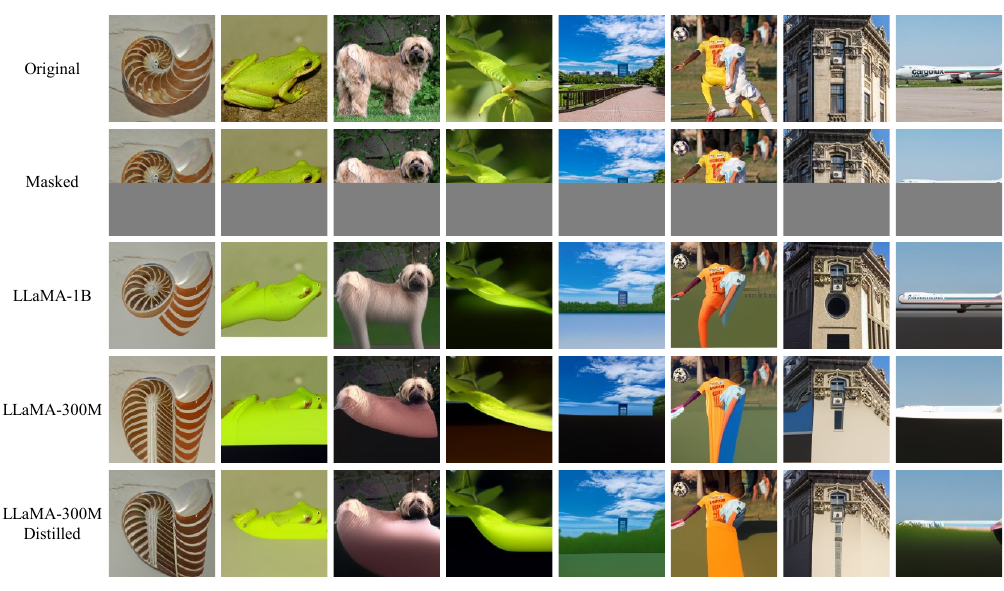}}
    \caption{Visualization of image inpainting results. Input of the top half of each image to LVMs to generate the inpainted bottom half.}
    \label{fig:supp:inpaint}
  \end{center}
  \vskip -0.2in
\end{figure}

Figure~\ref{fig:supp:inpaint} illustrates the visualization of single-image inpainting.
For each input sequence, comprised of 256 tokens representing a single image, the last 128 tokens are removed. 
Pretrained LVMs are then employed to predict the removed tokens.
Subsequently, the generated tokens are concatenated with the initial 128 tokens, and the VQGAN decoder is utilized to obtain the inpainted image.

%%%%%%%%%%%%%%%%%%%%%%%%%%%%%%%%%%%%%%%%%%%%%%%%%%%%%%%%%%%%%%%%%%%%%%%%%%%%%%%
%%%%%%%%%%%%%%%%%%%%%%%%%%%%%%%%%%%%%%%%%%%%%%%%%%%%%%%%%%%%%%%%%%%%%%%%%%%%%%%

\end{document}